\documentclass[10pt,twocolumn,letterpaper]{article}

\usepackage{cvpr}
\usepackage{times}
\usepackage{epsfig}
\usepackage{graphicx}
\usepackage{amsmath}
\usepackage{amssymb}
\usepackage{multirow}

\usepackage[pagebackref=true,breaklinks=true,letterpaper=true,colorlinks,bookmarks=false]{hyperref}

\cvprfinalcopy 

\def\cvprPaperID{2328} 

\ifcvprfinal\fi
\begin{document}

\title{Learning Spatial Regularization with Image-level Supervisions \\ 
	for Multi-label Image Classification}

\author{Feng Zhu$^{1,2}$,\ \ \ \ Hongsheng Li$^2$,\ \ \ \ Wanli Ouyang$^{2,3}$,\ \ \ \ Nenghai Yu$^1$,\ \ \ \ Xiaogang Wang$^2$\\
$^1$University of Science and Technology of China,~~~~$^3$University of Sydney\\
$^2$Department of Electronic Engineering, The Chinese University of Hong Kong\\
{\tt\small zhufengx@mail.ustc.edu.cn, \{hsli,wlouyang,xgwang\}@ee.cuhk.edu.hk, ynh@ustc.edu.cn}
}

\maketitle

\begin{abstract}
Multi-label image classification is a fundamental but challenging task in computer vision. Great progress has been achieved by exploiting semantic relations between labels in recent years. However, conventional approaches are unable to model the underlying spatial relations between labels in multi-label images, because spatial annotations of the labels are generally not provided. In this paper, we propose a unified deep neural network that exploits both semantic and spatial relations between labels with only image-level supervisions. Given a multi-label image, our proposed Spatial Regularization Network (SRN) generates attention maps for all labels and captures the underlying relations between them via learnable convolutions. By aggregating the regularized classification results with original results by a ResNet-101 network, the classification performance can be consistently improved. The whole deep neural network is trained end-to-end with only image-level annotations, thus requires no additional efforts on image annotations. Extensive evaluations on 3 public datasets with different types of labels show that our approach significantly outperforms state-of-the-arts and has strong generalization capability. Analysis of the learned SRN model demonstrates that it can effectively capture both semantic and spatial relations of labels for improving classification performance.
\end{abstract}
\vspace{-0.3cm}

\section{Introduction}
Multi-label image classification is an important task in computer vision with various applications, such as scene recognition \cite{boutell2004learning,shao2015deeply,shao2016slicing}, multi-object recognition \cite{lin2014microsoft,kang2016object,kang2016t}, human attribute recognition \cite{li2016human}, \etc. Compared to single-label image classification \cite{deng2009imagenet, fei2007learning, griffin2007caltech}, which has been extensively studied, multi-label problem is more practical and challenging, as real-world images are usually associated with multiple labels, such as objects or attributes. 

Binary relevance method \cite{tsoumakas2006multi} is an easy way to extend single-label algorithms to solve multi-label classification, which simply trains one binary classifier for each label. Various loss functions have been investigated in \cite{gong2013deep}. To cope with the problem that labels may relate to different visual regions over the whole image, proposal-based approaches \cite{wei2014cnn} are proposed to transform multi-label classification problem into multiple single-label classification tasks. However, these modifications of existing single-label algorithms ignored semantic relations of labels.

\begin{figure}[t]
	\begin{center}
		\includegraphics[width=0.9\linewidth]{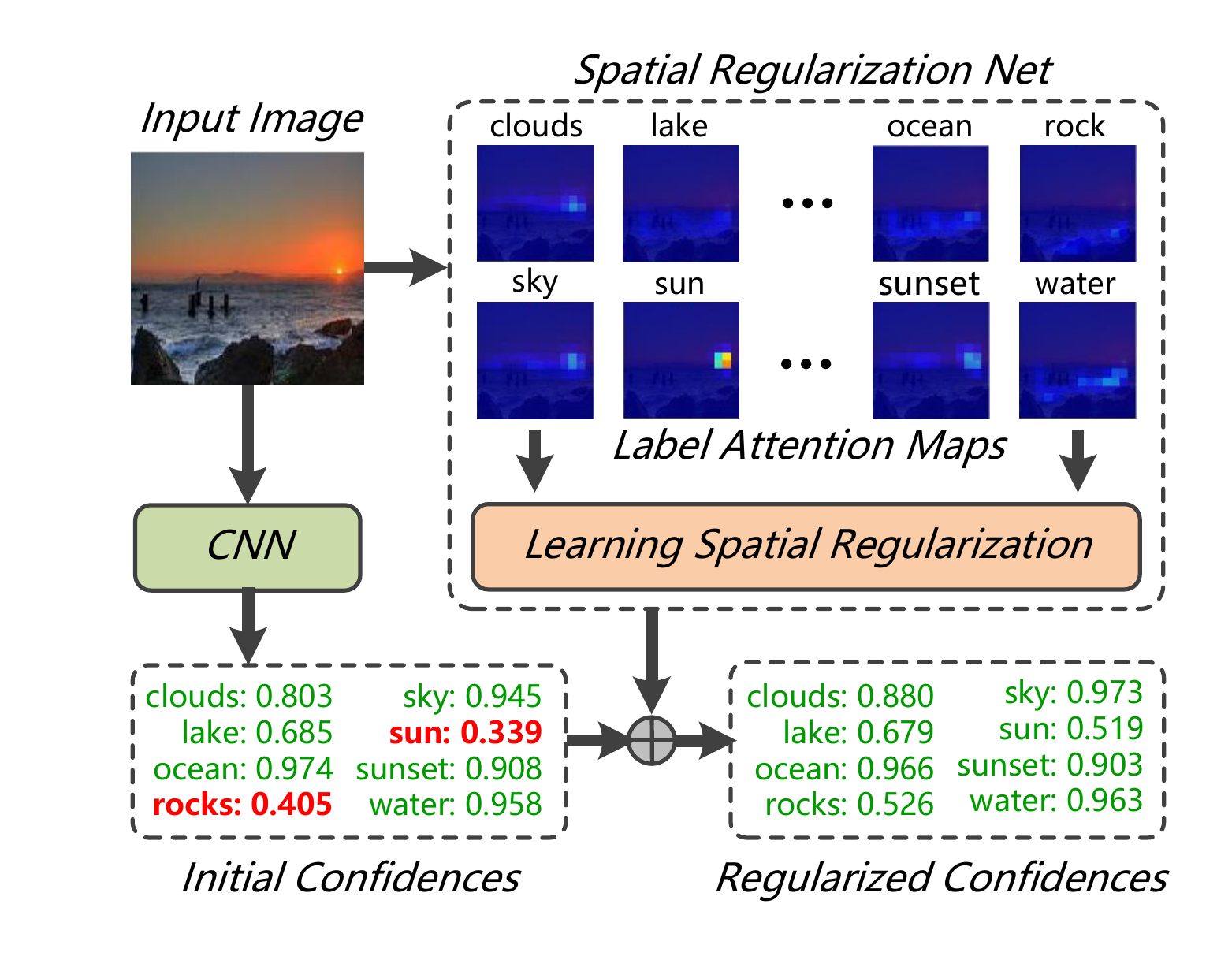}
	\end{center}
	\vspace{-0.1cm}
	\caption{Illustration of using our proposed Spatial Regularization Net (SRN) for improving multi-label image classification. The SRN learns semantic and spatial label relations from label attention maps with only image-level supervisions.}
	\label{fig:motivation}
	\vspace{-0.2cm}
\end{figure}

Recent progress on multi-label image classification mainly focused on capturing semantic relations between labels. Such relations or dependency can be modeled by probabilistic graphical models \cite{li2014multi, li2016conditional}, structured inference neural network \cite{hu2016learning}, or Recurrent Neural Networks (RNNs) \cite{wang2016cnn}. Despite the great improvements achieved by exploiting semantic relations, existing methods cannot capture spatial relations of labels, because their spatial locations are not annotated for training. In this paper, we propose to capture both semantic and spatial relations of labels by a Spatial Regularization Network in a unified framework (Figure \ref{fig:motivation}), which can be trained end-to-end with only image-level supervisions, thus requires no additional annotations. 

Deep Convolution Neural Networks (CNNs) \cite{krizhevsky2012imagenet,szegedy2015going,simonyan2014very,he2015deep} have achieved great success on single-label image classification in recent years. Because of their strong capability in learning discriminative features, deep CNN models pre-trained on large datasets can be easily transferred to solve other tasks and boost their performance. However, the feature representations might not be optimal for images with multiple labels, since a ground truth label might semantically relate to only a small region of the image. The diverse and complex contents in multi-label images make it difficult to learn effective feature representations and classifiers. 

Inspired by recent success of attention mechanism in many vision tasks \cite{xu2015show, yang2016stacked, hong2016learning}, we propose a deep neural network for multi-label classification, which consists of a sub-network, Spatial Regularization Net (SRN), to learn spatial regularizations between labels with only image-level supervisions. The SRN learns an attention map for each label, which associates related image regions to each label. By performing learnable convolutions on the attention maps of all labels, the SRN captures the underlying semantic and spatial relations between labels and act as spatial regularizations for multi-label classification.

The contribution of this paper is as follows. 1) We propose an end-to-end deep neural network for multi-label image classification, which exploits both semantic and spatial relations of labels by training learnable convolutions on the attention maps of labels. Such relations are learned with only image-level supervisions. Investigation and visualization of learned models demonstrate that our model can effectively capture semantic and spatial relations of labels. 2) Our proposed algorithm has great generalization capability and works well on data with different types of labels. We comprehensively evaluate our method on 3 publicly available datasets, NUS-WIDE \cite{chua2009nus} (81 concept labels), MS-COCO \cite{lin2014microsoft} (80 object labels), and WIDER-Attribute \cite{li2016human} (14 human attribute labels), showing significant improvements over state-of-the-art approaches.

\section{Related Work}

Multi-label classification has applications in many areas, such as document topic categorization \cite{schapire2000boostexter, godbole2004discriminative}, music annotation and retrieval \cite{turnbull2008semantic}, scene recognition \cite{boutell2004learning}, and gene functional analysis \cite{barutcuoglu2006hierarchical}. Comprehensive reviews for general multi-label classification methods can be found in \cite{tsoumakas2006multi, zhang2014review}. In this work, we focus on multi-label image classification methods with deep learning techniques.

A simple way of adapting existing single-label methods to multi-label is to learn an independent classifier for each label \cite{tsoumakas2006multi}. Recent success of deeply-learned features \cite{krizhevsky2012imagenet} for single-label image classification have boosted the accuracy of multi-label classification. Based on such deep features, Gong \etal \cite{gong2013deep} evaluated various loss functions and found that weighted approximate ranking loss worked best with CNNs. Proposal-based approaches showed promising performance in object detection \cite{girshick2015fast}. Similar ideas have also been explored for multi-label image classification. Wei \etal \cite{wei2014cnn} converted multi-label problems into a set of multi-class problems over region proposals. Classification results for the whole images were obtained by max-pooling label confidences over all proposals. Yang \etal \cite{yangexploit} treated images as a bag of instances/proposals, and solved a multi-instance learning problem. The above approaches ignored label relations in multi-label images.

Approaches that learn to capture label relations were also proposed. Read \etal \cite{read2011classifier} extended the binary relevance method by training a chain of binary classifiers, where each classifier makes predictions based on both image features and previously predicted labels. A more common way of modeling label relations is to use probabilistic graphical models \cite{koller2009probabilistic}. There were also methods on determining structures of the label relation graphs. Xue \etal \cite{xue2011correlative} directly thresholded the label correlation matrix to obtain the label structure. Li \etal \cite{li2014multi} used a maximum spanning tree over mutual information matrix of labels to create the graph. Li \etal \cite{li2016conditional} proposed to learn image-dependent conditional label structures base on Graphical Lasso framework \cite{ravikumar2010high}. Recently, deep neural networks have also been explored for learning label relations. Hu \etal \cite{hu2016learning} proposed a structured inference neural network that transfers predictions across multiple concept layers. Wang \etal \cite{wang2016cnn} treated multi-label classification as a sequential prediction problem, and solved label dependency by Recurrent Neural Networks (RNN). Although classification accuracy has been greatly improved by learning semantic relations of labels, the above mentioned approaches fail to explore the underlying spatial relations between labels.

\begin{figure*}
	\begin{center}
		\includegraphics[width=0.95\linewidth]{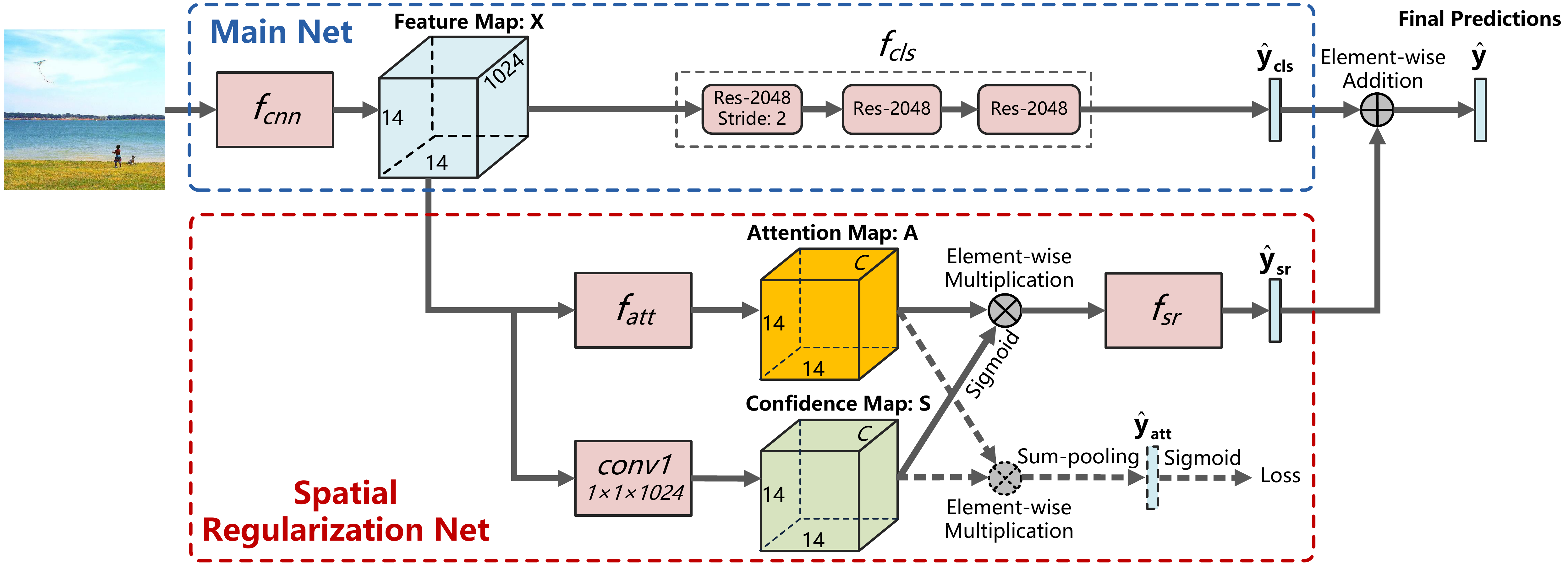}
	\end{center}
	\caption{Overall framework of our approach. (Top) The main net follows the structure of ResNet-101 and learns one independent classifier for each label. ``Res-2048'' stands for one ResNet building block with 2048 output channels. (Bottom) The proposed SRN captures spatial and semantic relations of labels with attention mechanism. Dashed lines indicate weakly-supervised pre-training for attention maps.}
	\label{fig:framework}
\end{figure*}

Attention mechanism was proven to be beneficial in many vision tasks, such as visual tracking \cite{bazzani2011learning}, object recognition \cite{mnih2014recurrent,ba2015multiple}, image captioning \cite{xu2015show}, image question answering \cite{yang2016stacked}, and segmentation \cite{hong2016learning}. The spatial attention mechanism adaptively focuses on related regions of the image when the deep networks are trained with spatially-related labels. In this paper, we utilize attention mechanism for improving multi-label image classification, which captures the underlying spatial relations of labels and provides spatial regularization for the final classification results.

\section{Methodology}
\label{sec:method}

We propose a deep neural network for multi-label classification, which utilizes image-level supervisions for learning spatial regularizations on multiple labels. The overall framework of our approach is shown in Figure \ref{fig:framework}. The main net has the same network structure as ResNet-101 \cite{he2015deep}. The proposed Spatial Regularization Net (SRN) takes visual features from the main net as inputs and learns to regularize spatial relations between labels. Such relations are exploited based on the learned attention maps for the multiple labels. Label confidences from both main net and SRN are aggregated to generate final classification confidences. The whole network is a unified framework and is trained in an end-to-end manner.

\subsection{Main Net for Multi-label Classification}
\label{sec:preliminary}

The main net follows the structure of ResNet-101 \cite{he2015deep} which is composed of repetitive building blocks with different output dimensions. Specifically, the block structure proposed in \cite{he2016identity} is adopted. The $14\times 14$ feature map (for $224\times 224$ input images) from layer ``\textit{res4b22\_relu}'' of the main net is used as input for SRN, which is of sufficient resolution to learn spatial regularizations in our experiments.

Let $\mathbf{I}$ denote an input image with ground-truth labels $\mathbf{y}=[y^1,y^2,...,y^C]^T$, where $y^l$ is a binary indicator. $y^l=1$ if image $\mathbf{I}$ is tagged with label $l$, and $y^l=0$ otherwise. $C$ is the number of all possible labels in the dataset. The main net conducts binary classification for each of the $C$ labels, 
\begin{equation}
\mathbf{X}=f_{cnn}(\mathbf{I};\theta_{cnn}),~~\mathbf{X} \in \mathbb{R}^{14\times 14\times 1024},
\end{equation}
\begin{equation}
\mathbf{\hat{y}}_{cls}=f_{cls}(\mathbf{X};\theta_{cls}),~~\mathbf{\hat{y}}_{cls} \in \mathbb{R}^C,
\end{equation}
where $\mathbf{X}$ is the feature map from layer ``\textit{res4b22\_relu}'', $\mathbf{\hat{y}}_{cls}=[\hat{y}_{cls}^1\cdots,\hat{y}_{cls}^C]^T$ is predicted label confidences by the main net. Prediction errors of the main net is measured based on $\mathbf{\hat{y}}_{cls}$ and ground-truth labels $\mathbf{y}$.

The proposed SRN is composed of two successive sub-networks, where the first sub-network $f_{att}(\mathbf{X};\theta_{att})$ learns label attention maps with image-level supervisions (Section \ref{sec:attention_learning}), and the second sub-network $f_{sr}(\mathbf{U};\theta_{sr})$ captures spatial regularizations of labels (Section \ref{sec:spatial_regularization}) based on the learned label attention maps.

 \subsection{Label Attention from Image-level Supervisions}
\label{sec:attention_learning}

Multi-label image is composed of multiple image regions that are semantically related to different labels. Although the region locations are generally not provided by the image-level supervisions, when predicting one label's existence, it is desirable that more attention is paid to the related regions. In our work, our neural network learn to predict such related image regions for each label with image-level supervisions using the attention mechanism. The learned attention maps could then be used to learn spatial regularizations for the labels.

Given input visual features $\mathbf{X} \in \mathbb{R}^{14\times 14\times 1024}$ from layer ``\textit{res4b22\_relu}'' of the main net, we aim to automatically generate label attention values for each individual labels,
\begin{equation}
\mathbf{Z}=f_{att}(\mathbf{X};\theta_{att}),~~\mathbf{Z} \in \mathbb{R}^{14\times14\times C},
\end{equation}
where $\mathbf{Z}$ is the unnormalized label attention values by $f_{att}(\cdot)$ with each channel corresponding to one label. Following \cite{xu2015show}, $\mathbf{Z}$ is spatially normalized with the softmax function to obtain the final label attention maps $\mathbf{A}$, 
\begin{equation}
a_{i,j}^l=\dfrac{\text{exp}~(z_{i,j}^l)}{\sum_{i,j}\text{exp}~(z_{i,j}^l)},\
~~\mathbf{A} \in \mathbb{R}^{14\times14\times C},
\end{equation}
where $z_{i,j}^l$ and $a_{i,j}^l$ represent the unormalized and normalized attention values at $(i,j)$ for label $l$.
Intuitively, if label $l$ is tagged to the input image, the image regions related to it should be assigned with higher attention values. The attention estimator $f_{att}(\cdot)$ is modeled as 3 convolution layers with 512 kernels of $1\times 1$, 512 kernels of $3\times 3$, and $C$ kernels of $1\times1$, respectively. The ReLU nonlinearity operations are performed following the first two convolution layers. 

Since ground-truth annotations of attention maps are not available, $f_{att}(\mathbf{X};\theta_{att})$ is learned with only image-level multi-label supervisions.
Let $\mathbf{x}_{i,j} \in \mathbb{R}^{1024}$ denote the visual feature vector at location $(i,j)$ of ${\bf X}$. In the original ResNet, the visual features is averaged across all spatial locations for classification as $\frac{1}{7\times7}\sum_{i,j}\mathbf{x}_{i,j}$. Since we expect the attention map $\mathbf{A}^l$ for each label $l$ to have higher values at the label-related regions, and $\sum_{i,j} a_{i,j}^l=1$ for all $l$, the attention maps could be used to weightedly average the visual features $\mathbf{X}$ for each label $l$ as,
\begin{align}
{\bf v}^l=\sum_{i,j}\mathbf{x}_{i,j} a_{i,j}^l,~~\mathbf{v}^l \in \mathbb{R}^{1024}.
\end{align}
Compared with the original averaged visual features shared by all labels, the weightedly-averaged visual feature vector ${\bf v}^l$ is more related to image regions corresponding to label $l$. Each such feature vector is then used to learn a linear classifier for estimating label $l$'s confidence,
\begin{align}
\hat{y}^l_{att} = W^l{\bf v}^l + b^l,
\label{eq:att_confidence}
\end{align}
where $W_l$ and $b_l$ are classifier parameters for label $l$. For all labels, ${\bf \hat{y}}_{att}= [\hat{y}^1_{att}, \cdots, \hat{y}^C_{att}]^T$. Using only the image-level supervisions ${\bf y}$ for training, the attention estimator parameters are learned by minimizing the cross-entropy loss between ${\bf \hat{y}}_{att}$ and ${\bf y}$ (see the dashed lines in Figure \ref{fig:framework}).

\begin{figure}[t]
	\begin{center}
		\includegraphics[width=0.9\linewidth]{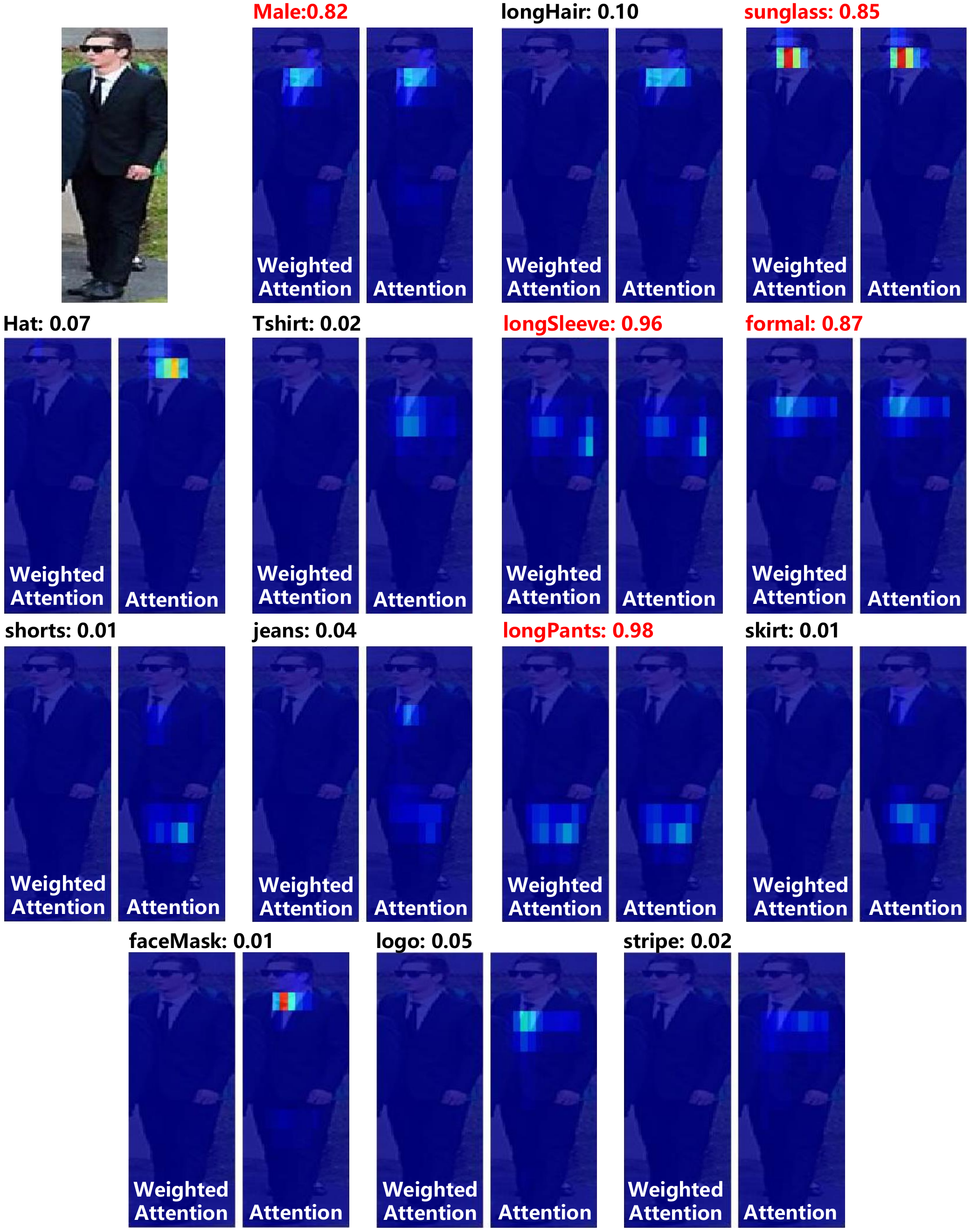}
	\end{center}
	\vspace{-0.3cm}
	\caption{Examples of learned attention maps from WIDER-Attribute dataset. Labels in red are ground-truth labels. ``Weighted Attention'' is the attention map weighted by corresponding label confidence (Eq. (\ref{equ:att_score_map})).}
	\label{fig:learned_attention}
\end{figure}

The attention estimator network $f_{att}(\cdot)$ can effectively learn attention maps for each label. Learned attention maps for an image are illustrated in Figure \ref{fig:learned_attention}. It shows that the weakly-supervised attention model could effectively capture related visual regions for each label. For example, \textit{``sunglass''} focuses on the face region, while \textit{``longPants''} pays more attention to legs. The negative labels also focus on reasonable regions, for example, \textit{``Hat''} tries to find hat in the region of head.

For efficient learning of the attention maps, recall that we have $\sum_{i,j}a_{i,j}^l=1$, and Eq. (\ref{eq:att_confidence}) can be rewritten as
\begin{align}
\hat{y}^l_{att} = \sum_{i,j}a_{i,j}^l (W^l \mathbf{x}_{i,j} + b^l).
\label{equ:score_gate}
\end{align}
This equation can be viewed as applying label-specific linear classifier at every location of the feature map $\mathbf{X}$, and then spatially aggregating label confidences based on attention maps. In our implementation, the linear classifiers are modeled as a convolution layer with $C$ kernels of size $1\times1$ (\textit{``conv1''} in Figure \ref{fig:framework}). The output of this layer is a confidence map $~\mathbf{S} \in \mathbb{R}^{14\times 14\times C}$, where its $l$th channel is ${\bf S}^l = W^l*\mathbf{X} + b^l$, with $*$ denoting convolution operation. The label attention map $\mathbf{A}$ and confidence map $\mathbf{S}$ are element-wisely multiplied, and then spatially sum-pooled to obtain the label confidence vector $\mathbf{\hat{y}}_{att}$. This formulation leads to an easy-to-implement network for learning label attentions, and generates confidence maps for weighting the attention maps in SRN.

\subsection{Spatial Regularizations from Attention Maps}
\label{sec:spatial_regularization}

Label attention maps encode rich spatial information of labels. They can be used to generate more robust spatial regularizations for labels. However, the attention map for each label always sum up to $1$ (see Figure \ref{fig:learned_attention}), which may highlight wrong locations. Learning from label-not-existing attention maps might lead to wrong spatial regularizations. Therefore, we propose to learn spatial regularizations from weighted attention maps $\mathbf{U} \in \mathbb{R}^{14\times 14\times C}$,
\begin{align}
\mathbf{U}&=\sigma(\mathbf{S})\circ \mathbf{A}, 
\label{equ:att_score_map}
\end{align}
where $\sigma(x)=1/(1+e^{-x})$ is the sigmoid function that converts label confidences $\mathbf{S}$ to the range $[0,1]$, and $\circ$ indicates element-wise multiplication. The weighted attention maps $\mathbf{U}$ encode both local confidences of attention and global visibility of each label, as shown in Figure \ref{fig:learned_attention}.

Given the weighted attention maps ${\bf U}$, a label regularization function is required to estimate the label confidences based on label spatial information from ${\bf U}$,
\begin{align}
\mathbf{\hat{y}}_{sr}=f_{sr}(\mathbf{U};\theta_{sr}),~~\mathbf{\hat{y}}_{sr} \in \mathbb{R}^C,
\label{equ:sr}
\end{align}
where $\mathbf{\hat{y}}_{sr}=[\hat{y}_{sr}^1,\hat{y}_{sr}^2,...,\hat{y}_{sr}^C]^T$ is predicted label confidences by the label regularization function.

Since the weighted attention maps for all labels are spatially aligned, it is easy to capture their relative relations with stacked convolution operations. The convolutions should have large enough receptive fields to capture the complex spatial relations between the labels. However, a naive implementation might be problematic. If we only use one convolution layer with 2048 filters of size $14\times 14$, then the total number of additional parameters would be $0.4C$ million. For a dataset that has 80 different labels, the actual number of additional parameter would be $32$ million, In contrast, the original ResNet-101 only contains approximately $40$ million parameters. Such large number of additional parameters would make the network difficult to train.

\begin{figure}[t]
	\begin{center}
		\includegraphics[width=0.88\linewidth]{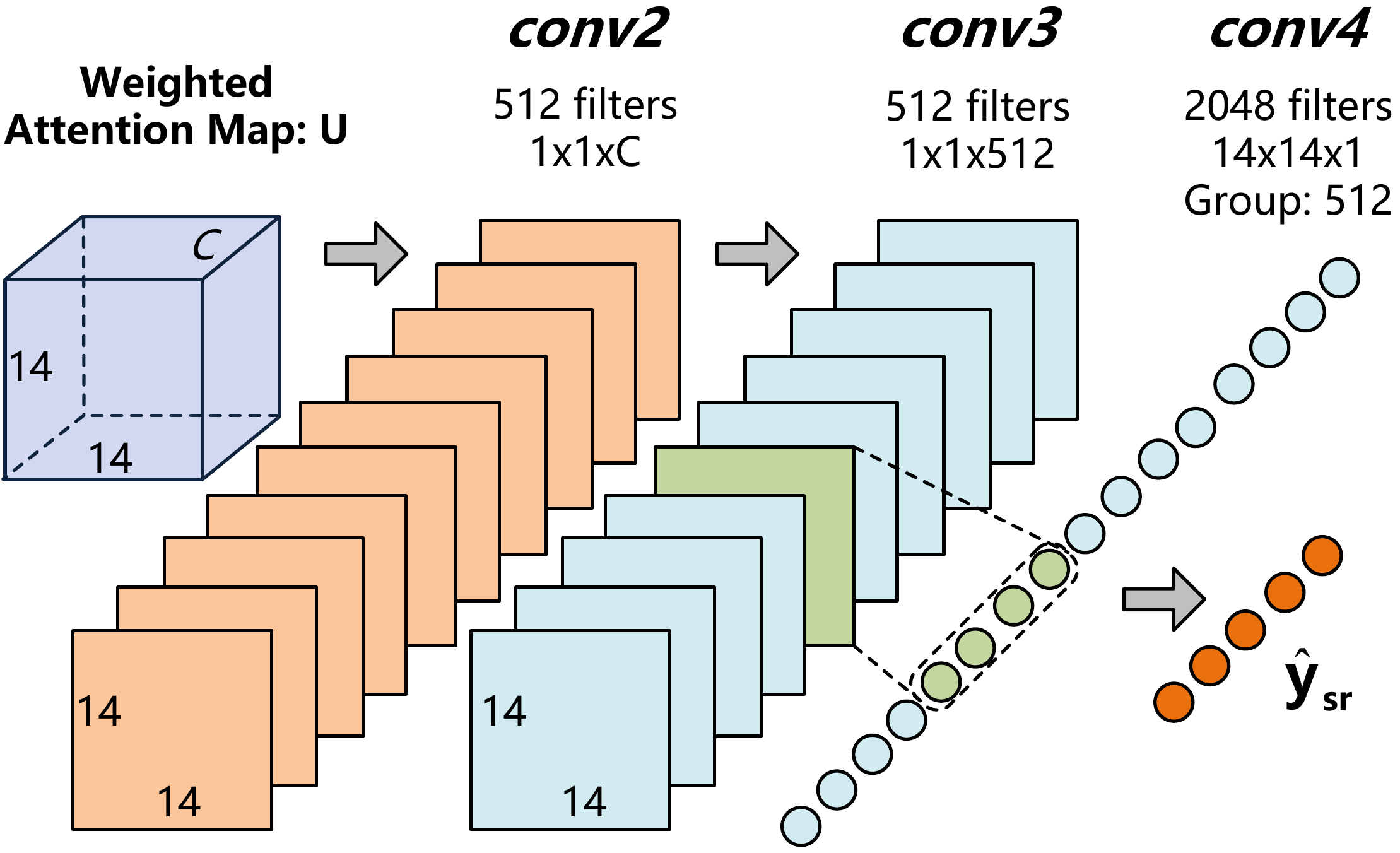}
	\end{center}
	\vspace{-0.3cm}
	\caption{Detailed network structure of $f_{sr}(\cdot)$ for learning spatial regularizations from weighted attention maps. The first two layers (\textit{``conv2''} and \textit{``conv3''}) are convolution layers with multi-channel filters, while, \textit{``conv4''} is composed of single-channel filters. Every 4 filters convolve with the same feature channel by \textit{``conv3''} to limit the parameter size.}
	\label{fig:srn}
\end{figure}

We propose to decouple semantic relation learning and spatial relation learning in different convolution layers. The intuition is that one label may only semantically relate to a small number of other labels, and measuring spatial relations with those unrelated attention maps is unnecessary. $f_{sr}(\mathbf{U};\theta_{sr})$ is implemented as three convolution layers with ReLU nonlinearity followed by one fully-connected layer as shown in Figure \ref{fig:srn}. The first two layers capture semantic relations of labels with 2 layers of $1\times1$ convolutions, and the third layer explores spatial relations using 2048 $14\times14$ kernels. The filters of the third convolution layer are grouped, with each group of 4 kernels corresponding to one feature channel of the input feature map. The 4 kernels in each group convolve the same feature channel independently. Different kernels in one group capture different spatial relations of semantically related labels. Experimental results show that the proposed SRN provides effective regularization to the classification results based on semantic and spatial relations of labels, with only about 6 million additional parameters.

\subsection{Overall Network and Training Scheme}
\label{sec:training}

The final label confidences are aggregation of the outputs of main net and SRN, $\mathbf{\hat{y}}=\alpha\mathbf{\hat{y}}_{cls}+(1-\alpha)\mathbf{\hat{y}}_{sr}$, where $\alpha$ is a weighting factor. Though the factor can also be learned, we fix $\alpha=0.5$ and do not observe performance drop. The whole network is trained with the cross-entropy loss with the ground truth labels ${\bf y}$,
\begin{equation}
F_{loss}(\mathbf{y},\mathbf{\hat{y}})=\sum_{l=1}^{C}y^l\log \sigma(\hat{y}^l)+(1-y^l)\log (1-\sigma(\hat{y}^l)).
\label{equ:loss_function}
\end{equation}

We train the network in multiple steps. First, we fine-tune only the main net on the target dataset, which is pre-trained on 1000-classification task of ImageNet dataset \cite{deng2009imagenet}.
Both $f_{cnn}(\mathbf{I};\theta_{cnn})$ and $f_{cls}(\mathbf{X};\theta_{cls})$ are learned with cross-entropy loss $F_{loss}(\mathbf{y},\mathbf{\hat{y}}_{cls})$. Secondly, we fix $f_{cnn}$ and $f_{cls}$, and focus on training $f_{att}(\mathbf{X};\theta_{att})$ and \textit{``conv1''} (see dashed lines in Figure \ref{fig:framework}) with loss $F_{loss}(\mathbf{y},\mathbf{\hat{y}}_{att})$. Thirdly, we train $f_{sr}(\mathbf{U};\theta_{sr})$ with cross-entropy loss $F_{loss}(\mathbf{y},\mathbf{\hat{y}}_{sr})$, by fixing all other sub-networks. Finally, the whole network is jointly fine-tuned with loss $F_{loss}(\mathbf{y},\mathbf{\hat{y}})+F_{loss}(\mathbf{y},\mathbf{\hat{y}}_{att})$.

Our deep neural network is implemented with Caffe library \cite{jia2014caffe}. To avoid over-fitting, we adopt image augmentation strategies suggested in \cite{MultiGPUCaffe2015}. The input images are first resized to $256\times256$, and then cropped at four corners and the center. The width and height of cropped patches are randomly chosen from the set $\{256,224,192,168,128\}$. Finally, the cropped images are all resized to $224\times224$. We employ stochastic gradient descend algorithm for training, with a batch size of 96, a momentum of 0.9, and weight decay of 0.0005. The initial learning rate is set as $10^{-3}$, and decreased to $1/10$ of the previous value whenever validation loss gets saturated, until $10^{-5}$. We train our model with 4 NVIDIA Titan X GPUs. For MS-COCO, training costs about 16 hours for all steps. For testing, we simply resize all images to $224\times224$ and conduct single-crop evaluation.

\begin{table*}
	\begin{center}
		\begin{tabular}{l|c|c|c|c|c|c|c||c|c|c|c|c|c}
			\hline
			\multirow{2}{*}{Method}      &                                           \multicolumn{7}{c||}{All}                                           &                                   \multicolumn{6}{c}{top-3}                                   \\ \cline{2-14}
			&      mAP    &     F1-C     &      P-C      &      R-C      &     F1-O     &      P-O      &      R-O      &     F1-C     &      P-C      &      R-C      &     F1-O     &      P-O      &      R-O      \\ \hline
			KNN \cite{chua2009nus}       &       -       &       -       &       -       &       -       &       -       &       -       &       -       &     24.3      &     32.6      &     19.3      &     47.6      &     42.9      &     53.4      \\
			WARP \cite{gong2013deep}     &       -       &       -       &       -       &       -       &       -       &       -       &       -       &     33.5      &     31.7      &     35.6      &     53.9      &     48.6      &     60.5      \\
			CNN-RNN \cite{wang2016cnn}   &       -       &       -       &       -       &       -       &       -       &       -       &       -       &     34.7      &     40.5      &     30.4      &     55.2      &     49.9      &     61.7      \\
			ResNet-101 \cite{he2015deep} &     59.8      &     55.7      &     65.8      &     51.9      &     72.5      &     75.9      &     69.5      &     47.0      &     46.9      &     56.8      &     61.7      &     55.8      &     69.1      \\
			ResNet-107                   &     59.5      &     55.6      &     65.4      &     52.2      &     72.6      &     75.5      &     70.0      &     46.9      &     46.7      &     56.8      &     61.8      &     55.9      &     69.2      \\
			ResNet-101-semantic          &     60.1      &     54.9      & \textbf{69.3} &     48.6      &     72.6      & \textbf{76.9} &     68.8      &     47.0      &     46.4      &     55.3      &     61.8      &     55.9      &     69.2      \\ \hline
			ResNet-SRN-att     &     61.8      &     56.9      &     67.5      &     52.5      &     73.2      &     76.5      &     70.1      &     47.7      &     47.4      &     57.7      &     \textbf{62.2}      &     \textbf{56.2}      &     \textbf{69.6}      \\
			ResNet-SRN                   & \textbf{62.0} & \textbf{58.5} &     65.2      & \textbf{55.8} & \textbf{73.4} &     75.5      & \textbf{71.5} & \textbf{48.9} & \textbf{48.2} & \textbf{58.9} & \textbf{62.2} & \textbf{56.2} & \textbf{69.6} \\ \hline
		\end{tabular}
	\end{center}
	\caption{Quantitative results by our proposed ResNet-SRN and compared methods on NUS-WIDE dataset. ``mAP'', ``F1-C'', ``P-C'', and ``R-C'' are evaluated for each class before averaging. ``F1-O'', ``P-O'', and ``R-O'' are averaged over all sample-label pairs.}
	\label{table:nus}
	\vspace{-0.1cm}
\end{table*}

\section{Experiments}
\label{sec:experiment}

Our approach is evaluated with three benchmark datasets with different types of labels:  NUS-WIDE \cite{chua2009nus} with 81 concept labels, MS-COCO \cite{lin2014microsoft} with 80 object labels, and WIDER-Attribute \cite{li2016human} with 14 human attribute labels. Experimental results show that our approach significantly outperforms state-of-the-arts on all the three datasets \footnote{Code and trained models available at \url{https://github.com/zhufengx/SRN_multilabel}.}, and has strong generalization capability to different types of labels. Analysis of the learned deep models demonstrates that our proposed approach can effectively capture both semantic and spatial relations of labels.

\subsection{Evaluation Metrics and Compared Methods}
\label{sec:metric}

\textbf{Evaluation Metrics}. A comprehensive study of evaluation metrics for multi-label classification is presented in \cite{wu2016unified}. We employ macro/micro precision, macro/micro recall, macro/micro F1-measure, and Mean Average Precision (mAP) for performance comparison. For precision/recall/F1-measure, if the estimated label confidences for any label are greater than 0.5, the labels are predicted as positive. Macro precision (denoted as ``P-C'') is evaluated by averaging per-class precisions, while micro precision (denoted as ``P-O'') is an overall measure which counts true predictions for all images over all classes. Similarly, we can also evaluate macro/micro recall (``R-C''/``R-O'') and macro/micro F1-measure (``F1-C''/``F1-O''). Mean Average Precision is the mean value of per-class average precisions. The above metrics do not require a fixed number of labels for each image. Generally, the mAP, F1-C and F1-O are of more importance.
To fairly compare with state-of-the-arts, we also evaluate precision, recall and F1-measure under the constraints that each image is predicted with top-3 labels. To obtain such top-3 labels in our approach, the 3 labels with highest confidences are obtained for each image even if their confidence values are lower than 0.5. However, we argue that outputting a variable number of labels for each image is more practical for real-world applications. Therefore, we report both our results with and without the top-3 label constraint.

\textbf{Compared Methods}. For NUS-WIDE and MS-COCO datasets, we compare with state-of-the-art methods on the datasets including CNN-RNN \cite{wang2016cnn}, WARP \cite{gong2013deep}, and KNN \cite{chua2009nus}. CNN-RNN explored semantic relations of labels, while other methods did not. For WIDER-Attribute dataset, RCNN \cite{girshick2015fast}, R*CNN \cite{gkioxari2015contextual}, and DHC \cite{li2016human} are compared. Both R*CNN and DHC explored spatial context surrounding human bounding boxes. For our approach (denoted as ``ResNet-SRN''), one variant is also explored, which learns spatial regularizations from unweighted attention maps ${\bf A}$ instead of ${\bf U}$ to evaluate the necessity of weighting the attention maps. It is denoted as ``ResNet-SRN-att''.

We also design three baseline methods to further validate the effectiveness of our proposed Spatial Regularization Net. The first baseline is the original ResNet-101 (denoted as ``ResNet-101'') fine-tuned on each of the datasets. For the second baseline, since the proposed SRN has about 6 million additional parameters compared with ResNet-101, which is approximately equal to two ResNet building blocks with 2048 output feature channels, we add two such residual blocks following the last block of ResNet-101 (the layer \textit{``res5c\_relu''}) to create a ``ResNet-107'' model. For the third baseline, we investigate learning semantic relations of labels based on initial label confidences by ResNet-101. The initial confidences are concatenated with the visual features from the \textit{``pool5''} layer to encode label relations. Two 2048-neuron and one $C$-neuron fully-connected layers try to capture label semantic relations from the concatenated features to generate final label confidences. We refer this model as ``ResNet-101-semantic'' in our experiments.

\begin{table*}
	\begin{center}
		\begin{tabular}{l|c|c|c|c|c|c|c||c|c|c|c|c|c}
			\hline
			\multirow{2}{*}{Method}      &           \multicolumn{7}{c||}{All}            &         \multicolumn{6}{c}{top-3}         \\ \cline{2-14}
			& mAP & F1-C & P-C  & R-C  & F1-O & P-O  & R-O  & F1-C & P-C  & R-C  & F1-O & P-O  & R-O  \\ \hline
			WARP \cite{gong2013deep}     &  -   &  -   &  -   &  -   &  -   &  -   &  -   & 55.7 & 59.3 & 52.5 & 60.7 & 59.8 & 61.4 \\
			CNN-RNN \cite{wang2016cnn}   &  -   &  -   &  -   &  -   &  -   &  -   &  -   & 60.4 & 66.0 & 55.6 & 67.8 & 69.2 & \textbf{66.4} \\
			ResNet-101 \cite{he2015deep} & 75.2 & 69.5 & 80.8 & 63.4 & 74.4 & 82.2 & 68.0 & 65.9 & 84.3 & 57.4 & 71.7 & 86.5 & 61.3 \\
			ResNet-107                   & 75.4 & 69.7 & 80.9 & 63.7 & 74.5 & 82.1 & 68.2 & 66.1 & 84.4 & 57.6 & 71.8 & 86.4 & 61.4 \\
			ResNet-101-semantic           & 75.5 & 69.9 & 81.1 & 63.8 & 74.8 & 82.1 & 68.6 & 66.2 & 84.3 & 57.7 & 72.0 & 86.3 & 61.8 \\ \hline
			ResNet-SRN-att        &     76.1      &     70.0      &     81.2      &     63.3      &     75.0      &     \textbf{84.1}      &     67.7      &     66.3      &     \textbf{85.8}      &     57.5      &     72.1      &     \textbf{88.1}      &     61.1      \\
			ResNet-SRN                         & \textbf{77.1} & \textbf{71.2} & \textbf{81.6} & \textbf{65.4} & \textbf{75.8} & 82.7 & \textbf{69.9} & \textbf{67.4} & 85.2 & \textbf{58.8} & \textbf{72.9} & 87.4 & 62.5 \\ \hline
		\end{tabular}
	\end{center}
	\caption{Quantitative results by our proposed ResNet-SRN and compared methods on MS-COCO validation set. ``mAP'', ``F1-C'', ``P-C'', and ``R-C'' are evaluated for each class before averaging. ``F1-O'', ``P-O'', and ``R-O'' are averaged over all sample-label pairs.}
	\label{table:coco}
	\vspace{-0.1cm}
\end{table*}

\subsection{Experimental Results}

\textbf{NUS-WIDE} \cite{chua2009nus}. This dataset contains 269,648 images and associated tags from Flickr. The dataset is manually annotated by 81 concepts, with 2.4 concept labels per image on average. The concepts include events/activities (\eg, \textit{``swimming''}, \textit{``running''}), scene/location (\eg, \textit{``airport''}, \textit{``ocean''}), objects (\eg, \textit{``animal''}, \textit{``car''}). We trained our approach to predict the 81 concept labels. Official train/test split is used, \ie 161,789 images for training/validation, and 107,859 image for testing.

Experimental results on this dataset are shown in Table \ref{table:nus}. Our proposed ResNet-SRN and its variant ResNet-SRN-att outperform all state-of-the-arts and baseline models. With the advances of deep network structures, even our baseline ResNet-101 has achieved better performance than existing state-of-the-arts. It mainly results from the learning capability of ResNet-101 with deep learnable layers. When adding more layers to match the parameter size of our proposed SRN, ResNet-107 shows very close performance with ResNet-101, which suggests that the capacity of ResNet-101 is sufficient on NUS-WIDE, and adding more parameters does not lead to performance increase. Utilizing predicted labels as context (ResNet-101-semantic) dose not improve performance on this dataset. In contrast, by exploring spatial and semantic relations of labels, our proposed ResNet-SRN model outperforms all baseline methods by $\sim$ 2 percent. It indicates that learned spatial relations of labels provide good regularizations for multi-label image classification. The performance gain of our ResNet-SRN over ResNet-SRN-att suggests that the weighted attention map $\mathbf{U}$ is more informative for learning spatial regularizations than the unweighted attention map $\mathbf{A}$.

Forcing the algorithm to predict a fixed number ($k=3$ is proposed in state-of-the-art methods) of labels for each image may not fully reflect the algorithm's actual performance. When removing the constraint (Section \ref{sec:metric}), we can observe significant performance improvements (\eg, from 48.9 to 58.5 for the F1-C metric of ResNet-SRN).

\textbf{MS-COCO} \cite{lin2014microsoft}. This dataset is primarily built for object recognition task in the context of scene understanding. The training set is composed of 82,783 images, which contain common objects in the scenes. The objects are categorized into 80 classes, with about 2.9 object labels per image. Since the ground-truth labels of test set is not available, we evaluated all methods on the validation set (40,504 images). The number of labels for each image varies considerably on this MS-COCO. Following \cite{wang2016cnn}, when evaluating with top-3 label predictions, we filtered out labels with probability lower than 0.5 for each image, thus the image may return less than $k=3$ labels.

Quantitative results on MS-COCO are presented in Table \ref{table:coco}. The comparison results are similar to those on NUS-WIDE. Based on ResNet-101 network, all baseline models perform better than state-of-the-art approaches. ResNet-107 shows a minor improvement over ResNet-101. Due to more labels per image (3.5 labels on MS-COCO compared with 2.4 labels on NUS-WIDE), exploring label semantic relations by ResNet-101-semantic is helpful, but the improvement is limited (\eg, from 75.2 to 75.5 in terms of mAP). Both ResNet-SRN and ResNet-SRN-att show superior performance over baseline models, while the spatial regularizations learned from weighted attention maps perform better (\eg, ResNet-SRN boosts mAP to 77.1, as compared with 76.1 of ResNet-SRN-att).

\textbf{WIDER-Attribute} \cite{li2016human}. This dataset contains 13,789 images and 57,524 human bounding boxes. The task is to predict existence of 14 human attributes for each annotated person. Each image is also labeled by an event label from 30 event classes for context learning. For our approach, human is cropped from the full image based on bounding box annotations, and then used for training and testing. The training/validation and testing set contain 28,340 and 29,177 person, respectively. WIDER-Attribute also contains unspecified labels. We treated these unspecified labels as negative labels during training. Unspecified labels are excluded from evaluation in testing following the settings of \cite{li2016human}.

\begin{table}
	\begin{center}
		\begin{tabular}{l|c|c|c}
			\hline
			\multirow{2}{*}{Method}             &            \multicolumn{3}{c}{All}            \\ \cline{2-4}
			&      mAP      &     F1-C      &     F1-O      \\ \hline
			R-CNN \cite{girshick2015fast}       &     80.0      &       -       &       -       \\
			R*CNN \cite{gkioxari2015contextual} &     80.5      &       -       &       -       \\
			DHC \cite{li2016human}              &     81.3      &       -       &       -       \\
			ResNet-101 \cite{he2015deep}        &     85.0      &     74.7      &     80.4      \\
			ResNet-107                          &     85.0      &     74.8      &     80.6      \\
			ResNet-101-semantic                 &     85.1      &     74.8      &     80.5      \\ \hline
			ResNet-SRN-att				&	  85.4		& 	  74.9		&     80.8		\\
			ResNet-SRN                          & \textbf{86.2} & \textbf{75.9} & \textbf{81.3} \\ \hline
		\end{tabular}
	\end{center}
	\caption{Quantitative results by our proposed ResNet-SRN and compared methods on WIDER-Attribute dataset. ``mAP''and ``F1-C'' are evaluated for each class before averaging. ``F1-O'' is averaged over all sample-label pairs.}
	\label{table:wider}
	\vspace{-0.25cm}
\end{table}

Experimental results are shown in Table \ref{table:wider}. All ResNet models outperform state-of-the-arts, R-CNN \cite{girshick2015fast}, R*CNN \cite{gkioxari2015contextual}, and DHC \cite{li2016human}, and our proposed ResNet-SRN  performs best. It is important to note that R*CNN and DHC explore visual context surrounding the target human by taking full images and human bounding boxes as input. Event labels associated with each image are also used for training in DHC. In contrast, our approach and baselines only utilize cropped image patches without using the event labels. Nevertheless, the ResNet-SRN and ResNet-SRN-att show consistent improvement over state-of-the-arts and baseline methods. This result indicates that the proposed SRN could capture the spatial relations of human attributes with image-level supervisions, and these learned spatial regularizations could help predicting human attributes.

\subsection{Visualization and Analysis}

\begin{figure*}[t]
	\begin{center}
		\includegraphics[width=0.90\linewidth]{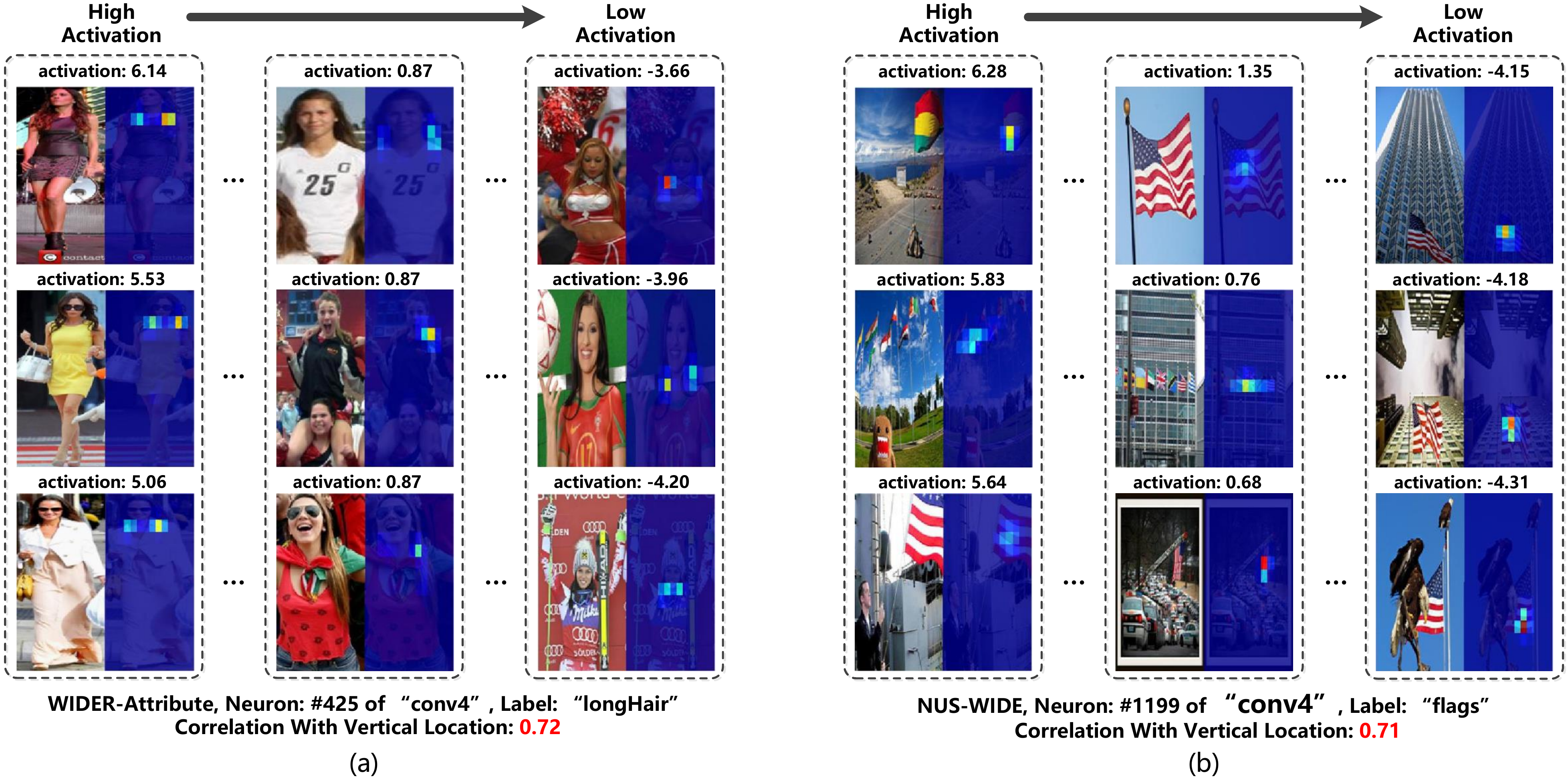}
		\vspace{-0.4cm}
	\end{center}
	\caption{Correlation between neuron activations and label locations. These two neurons are sensitive to the location variations of corresponding labels.}
	\label{fig:correlation_single}
	\vspace{-0.3cm}
\end{figure*}

The effectiveness of our approach has been quantitatively evaluated in Table \ref{table:nus}, \ref{table:coco}, and \ref{table:wider}, we visualize and analyze the learned neurons from the \textit{conv4} layer of our SRN to illustrate its capability of learning spatial regularizations for labels. We observe that the learned neurons capture two types of label spatial information. One type of neurons capture the spatial locations of individual labels, while the other type of neurons are only activated when several labels have specific relative position patterns.

We calculate correlations between learned neuron responses and label locations in images, and find some neuron highly correlates to individual label's spatial locations. In Figure \ref{fig:correlation_single}, we show two such examples. In (a), the response of neuron \#425 of \textit{``conv4''} in SRN highly correlates the vertical location of the label \textit{``longHair''} in WIDER-Attribute dataset. In (b) the activation of neuron \#1199 of \textit{``conv4''} highly correlates the vertical location of the label \textit{``flag''}. It demonstrates that the two neurons focuses on spatial locations of certain labels.

\begin{figure}[t]
	\begin{center}
		\includegraphics[width=0.78\linewidth]{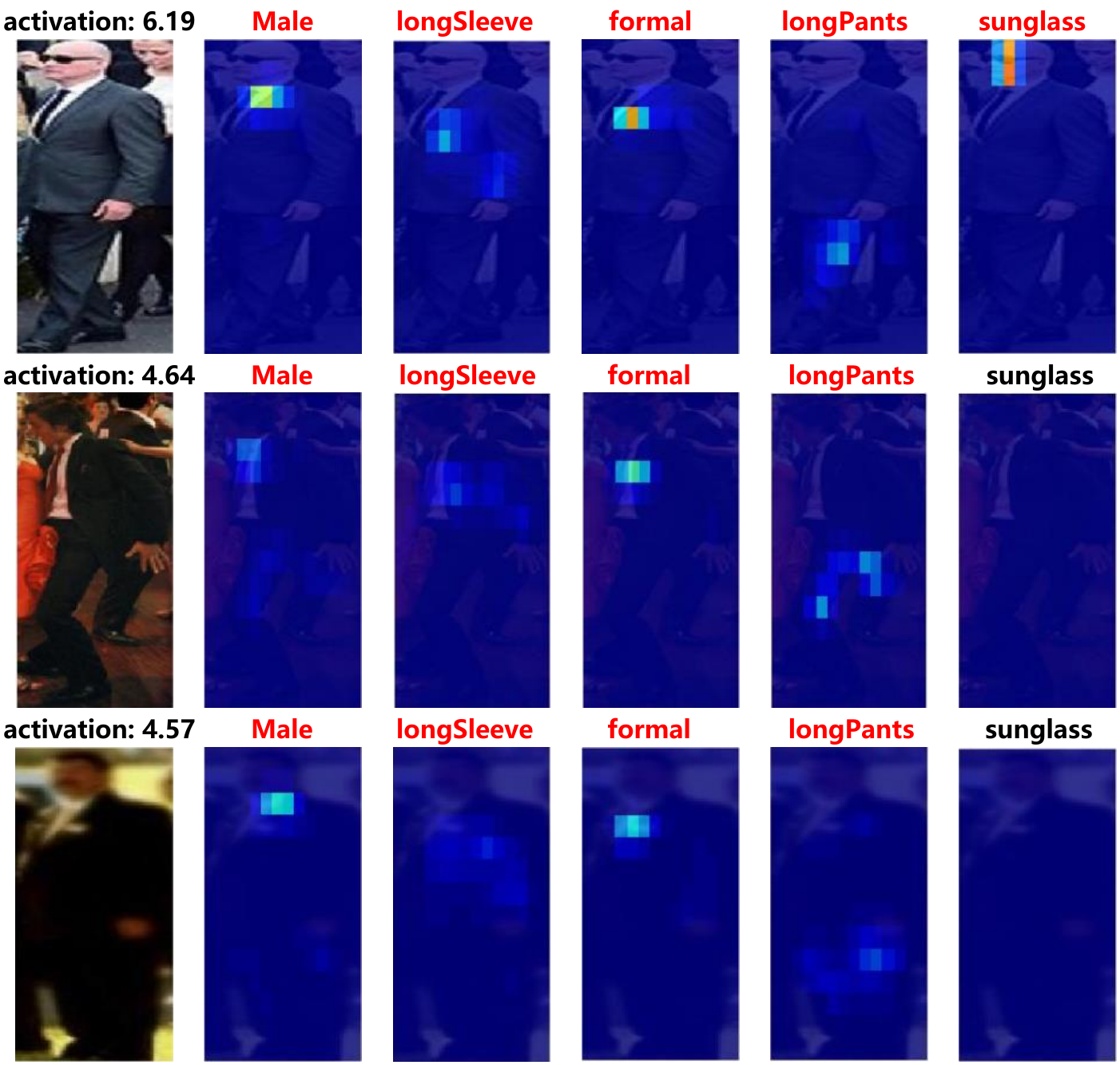}
	\end{center}
	\vspace{-0.3cm}
	\caption{Images with top-3 activations for neuron \#786 of \textit{``conv4''} from WIDER-Attribute dataset. True positive labels are marked in red. Strong spatial and semantic relations between the four labels (\textit{``Male''}, \textit{``longSleeve''}, \textit{``formal''}, \textit{``longPants''}) are captured by neuron.}
	\label{fig:semantic}
	\vspace{-0.4cm}
\end{figure}

In Figure \ref{fig:semantic}, we show three images from WIDER-Attribute dataset that have highest activations on neuron \#786 of \textit{``conv4''} in SRN. The images have common labels (\textit{``Male''}, \textit{``longSleeve''}, \textit{``formal''}, \textit{``longPants''}) with similar relative label positions. It suggests that this neuron is trained to capture semantic and spatial relations of the four labels, and favors specific relative positions between them.

\begin{figure}[t]
	\begin{center}
		\includegraphics[width=0.85\linewidth]{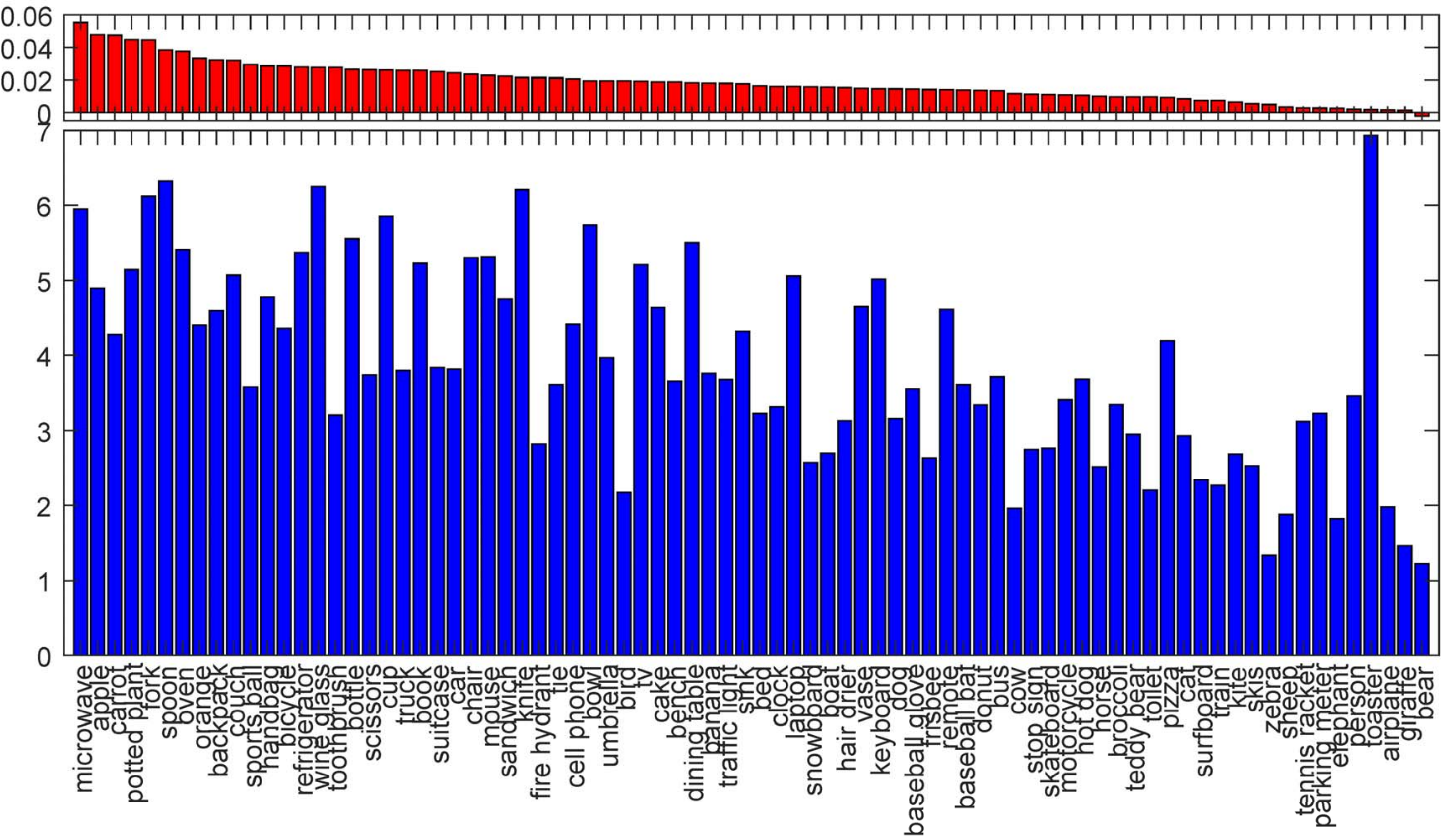}
	\end{center}
	\vspace{-0.3cm}
	\caption{Top: improvement in AP for each class in COCO. Bottom: average number of concurrent labels for true positive images of each class. All sorted according to improvements in AP.}
	\label{fig:APs}
	\vspace{-0.4cm}
\end{figure}

We also analyzed AP improvements for all classes in COCO. As shown in Figure \ref{fig:APs}, our approach is more effective for classes having more co-existing labels in the same images so that spatial relations can be better utilized to regularize the results. For the class \emph{toaster}, it was not improved much because of its limited number of training samples.

\vspace{-0.1cm}
\section{Conclusion}
\vspace{-0.1cm}
In this paper, we aim to improve multi-label image classification by exploring spatial relations of labels. This is achieved by learning attention maps for all labels with only image-level supervisions, and then capturing both semantic and spatial relations of labels base on weighted attention maps. Extensive evaluations on NUS-WIDE, MS-COCO, and WIDER-Attribute datasets show that our proposed Spatial Regularization Net significantly outperforms state-of-the-arts. Visualization of learned models also shows that our approach could effectively capture both semantic and spatial relations for labels.

\vspace{-0.1cm}
\section{Acknowledgment}
\vspace{-0.1cm}
This work is supported in part by National Natural Science Foundation of China under Grant 61371192, in part by SenseTime Group Limited, in part by the General Research Fund through the Research Grants Council of Hong Kong under Grants CUHK14213616, CUHK14206114, CUHK14205615, CUHK419412, CUHK14203015, CUHK14239816 and CUHK14207814, in part by the Hong Kong Innovation and Technology Support Programme Grant ITS/121/15FX, in part by the Ph.D. Program Foundation of China under Grant 20130185120039, and in part by the China Postdoctoral Science Foundation under Grant 2014M552339.

{\small
	\bibliographystyle{ieee}
	\bibliography{multilabel_cvpr17}

\begin{thebibliography}{10}\itemsep=-1pt

\bibitem{ba2015multiple}
J.~Ba, V.~Mnih, and K.~Kavukcuoglu.
\newblock Multiple object recognition with visual attention.
\newblock {\em ICLR}, 2015.

\bibitem{barutcuoglu2006hierarchical}
Z.~Barutcuoglu, R.~E. Schapire, and O.~G. Troyanskaya.
\newblock Hierarchical multi-label prediction of gene function.
\newblock {\em Bioinformatics}, 22(7):830--836, 2006.

\bibitem{bazzani2011learning}
L.~Bazzani, H.~Larochelle, V.~Murino, J.-a. Ting, and N.~D. Freitas.
\newblock Learning attentional policies for tracking and recognition in video
  with deep networks.
\newblock In {\em ICML}, 2011.

\bibitem{boutell2004learning}
M.~R. Boutell, J.~Luo, X.~Shen, and C.~M. Brown.
\newblock Learning multi-label scene classification.
\newblock {\em Pattern recognition}, 37(9):1757--1771, 2004.

\bibitem{chua2009nus}
T.-S. Chua, J.~Tang, R.~Hong, H.~Li, Z.~Luo, and Y.~Zheng.
\newblock Nus-wide: a real-world web image database from national university of
  singapore.
\newblock In {\em Proceedings of the ACM international conference on image and
  video retrieval}, 2009.

\bibitem{deng2009imagenet}
J.~Deng, W.~Dong, R.~Socher, L.-J. Li, K.~Li, and L.~Fei-Fei.
\newblock Imagenet: A large-scale hierarchical image database.
\newblock In {\em CVPR}, 2009.

\bibitem{fei2007learning}
L.~Fei-Fei, R.~Fergus, and P.~Perona.
\newblock Learning generative visual models from few training examples: An
  incremental bayesian approach tested on 101 object categories.
\newblock {\em Computer Vision and Image Understanding}, 106(1):59--70, 2007.

\bibitem{girshick2015fast}
R.~Girshick.
\newblock Fast r-cnn.
\newblock In {\em ICCV}, 2015.

\bibitem{gkioxari2015contextual}
G.~Gkioxari, R.~Girshick, and J.~Malik.
\newblock Contextual action recognition with r*cnn.
\newblock In {\em ICCV}, 2015.

\bibitem{godbole2004discriminative}
S.~Godbole and S.~Sarawagi.
\newblock Discriminative methods for multi-labeled classification.
\newblock In {\em Pacific-Asia Conference on Knowledge Discovery and Data
  Mining}, 2004.

\bibitem{gong2013deep}
Y.~Gong, Y.~Jia, T.~Leung, A.~Toshev, and S.~Ioffe.
\newblock Deep convolutional ranking for multilabel image annotation.
\newblock {\em ICLR}, 2014.

\bibitem{griffin2007caltech}
G.~Griffin, A.~Holub, and P.~Perona.
\newblock Caltech-256 object category dataset.
\newblock 2007.

\bibitem{he2015deep}
K.~He, X.~Zhang, S.~Ren, and J.~Sun.
\newblock Deep residual learning for image recognition.
\newblock {\em arXiv preprint arXiv:1512.03385}, 2015.

\bibitem{he2016identity}
K.~He, X.~Zhang, S.~Ren, and J.~Sun.
\newblock Identity mappings in deep residual networks.
\newblock In {\em ECCV}, pages 630--645, 2016.

\bibitem{hong2016learning}
S.~Hong, J.~Oh, B.~Han, and H.~Lee.
\newblock Learning transferrable knowledge for semantic segmentation with deep
  convolutional neural network.
\newblock {\em CVPR}, 2016.

\bibitem{hu2016learning}
H.~Hu, G.-T. Zhou, Z.~Deng, Z.~Liao, and G.~Mori.
\newblock Learning structured inference neural networks with label relations.
\newblock {\em CVPR}, 2016.

\bibitem{jia2014caffe}
Y.~Jia, E.~Shelhamer, J.~Donahue, S.~Karayev, J.~Long, R.~Girshick,
  S.~Guadarrama, and T.~Darrell.
\newblock Caffe: Convolutional architecture for fast feature embedding.
\newblock {\em arXiv preprint arXiv:1408.5093}, 2014.

\bibitem{kang2016t}
K.~Kang, H.~Li, J.~Yan, X.~Zeng, B.~Yang, T.~Xiao, C.~Zhang, Z.~Wang, R.~Wang,
  X.~Wang, et~al.
\newblock T-cnn: Tubelets with convolutional neural networks for object
  detection from videos.
\newblock {\em arXiv preprint arXiv:1604.02532}, 2016.

\bibitem{kang2016object}
K.~Kang, W.~Ouyang, H.~Li, and X.~Wang.
\newblock Object detection from video tubelets with convolutional neural
  networks.
\newblock In {\em Proceedings of the IEEE Conference on Computer Vision and
  Pattern Recognition}, pages 817--825, 2016.

\bibitem{koller2009probabilistic}
D.~Koller and N.~Friedman.
\newblock {\em Probabilistic Graphical Models: Principles and Techniques}.
\newblock MIT Press, 2009.

\bibitem{krizhevsky2012imagenet}
A.~Krizhevsky, I.~Sutskever, and G.~E. Hinton.
\newblock Imagenet classification with deep convolutional neural networks.
\newblock In {\em NIPS}, 2012.

\bibitem{li2016conditional}
Q.~Li, M.~Qiao, W.~Bian, and D.~Tao.
\newblock Conditional graphical lasso for multi-label image classification.
\newblock In {\em CVPR}, 2016.

\bibitem{li2014multi}
X.~Li, F.~Zhao, and Y.~Guo.
\newblock Multi-label image classification with a probabilistic label
  enhancement model.
\newblock {\em Proc. Uncertainty in Artificial Intell}, 2014.

\bibitem{li2016human}
Y.~Li, C.~Huang, C.~C. Loy, and X.~Tang.
\newblock Human attribute recognition by deep hierarchical contexts.
\newblock In {\em ECCV}, 2016.

\bibitem{lin2014microsoft}
T.-Y. Lin, M.~Maire, S.~Belongie, J.~Hays, P.~Perona, D.~Ramanan,
  P.~Doll{\'a}r, and C.~L. Zitnick.
\newblock Microsoft coco: Common objects in context.
\newblock In {\em ECCV}, 2014.

\bibitem{mnih2014recurrent}
V.~Mnih, N.~Heess, A.~Graves, et~al.
\newblock Recurrent models of visual attention.
\newblock In {\em NIPS}, 2014.

\bibitem{ravikumar2010high}
P.~Ravikumar, M.~J. Wainwright, J.~D. Lafferty, et~al.
\newblock High-dimensional ising model selection using ℓ1-regularized
  logistic regression.
\newblock {\em The Annals of Statistics}, 38(3):1287--1319, 2010.

\bibitem{read2011classifier}
J.~Read, B.~Pfahringer, G.~Holmes, and E.~Frank.
\newblock Classifier chains for multi-label classification.
\newblock {\em Machine learning}, 85(3):333--359, 2011.

\bibitem{schapire2000boostexter}
R.~E. Schapire and Y.~Singer.
\newblock Boostexter: A boosting-based system for text categorization.
\newblock {\em Machine learning}, 39(2-3):135--168, 2000.

\bibitem{shao2015deeply}
J.~Shao, K.~Kang, C.~Change~Loy, and X.~Wang.
\newblock Deeply learned attributes for crowded scene understanding.
\newblock In {\em Proceedings of the IEEE Conference on Computer Vision and
  Pattern Recognition}, pages 4657--4666, 2015.

\bibitem{shao2016slicing}
J.~Shao, C.-C. Loy, K.~Kang, and X.~Wang.
\newblock Slicing convolutional neural network for crowd video understanding.
\newblock In {\em Proceedings of the IEEE Conference on Computer Vision and
  Pattern Recognition}, pages 5620--5628, 2016.

\bibitem{simonyan2014very}
K.~Simonyan and A.~Zisserman.
\newblock Very deep convolutional networks for large-scale image recognition.
\newblock {\em arXiv preprint arXiv:1409.1556}, 2014.

\bibitem{szegedy2015going}
C.~Szegedy, W.~Liu, Y.~Jia, P.~Sermanet, S.~Reed, D.~Anguelov, D.~Erhan,
  V.~Vanhoucke, and A.~Rabinovich.
\newblock Going deeper with convolutions.
\newblock In {\em CVPR}, 2015.

\bibitem{tsoumakas2006multi}
G.~Tsoumakas and I.~Katakis.
\newblock Multi-label classification: An overview.
\newblock {\em International Journal of Data Warehousing and Mining}, 2007.

\bibitem{turnbull2008semantic}
D.~Turnbull, L.~Barrington, D.~Torres, and G.~Lanckriet.
\newblock Semantic annotation and retrieval of music and sound effects.
\newblock {\em IEEE Transactions on Audio, Speech, and Language Processing},
  16(2):467--476, 2008.

\bibitem{wang2016cnn}
J.~Wang, Y.~Yang, J.~Mao, Z.~Huang, C.~Huang, and W.~Xu.
\newblock Cnn-rnn: A unified framework for multi-label image classification.
\newblock {\em CVPR}, 2016.

\bibitem{MultiGPUCaffe2015}
L.~Wang, Y.~Xiong, Z.~Wang, and Y.~Qiao.
\newblock Towards good practices for very deep two-stream convnets.
\newblock {\em CoRR}, abs/1507.02159, 2015.

\bibitem{wei2014cnn}
Y.~Wei, W.~Xia, J.~Huang, B.~Ni, J.~Dong, Y.~Zhao, and S.~Yan.
\newblock Cnn: Single-label to multi-label.
\newblock {\em arXiv preprint arXiv:1406.5726}, 2014.

\bibitem{wu2016unified}
X.-Z. Wu and Z.-H. Zhou.
\newblock A unified view of multi-label performance measures.
\newblock {\em arXiv preprint arXiv:1609.00288}, 2016.

\bibitem{xu2015show}
K.~Xu, J.~Ba, R.~Kiros, K.~Cho, A.~Courville, R.~Salakhutdinov, R.~S. Zemel,
  and Y.~Bengio.
\newblock Show, attend and tell: Neural image caption generation with visual
  attention.
\newblock {\em ICML}, 2015.

\bibitem{xue2011correlative}
X.~Xue, W.~Zhang, J.~Zhang, B.~Wu, J.~Fan, and Y.~Lu.
\newblock Correlative multi-label multi-instance image annotation.
\newblock In {\em ICCV}, 2011.

\bibitem{yangexploit}
H.~Yang, J.~T. Zhou, Y.~Zhang, B.-B. Gao, J.~Wu, and J.~Cai.
\newblock Exploit bounding box annotations for multi-label object recognition.
\newblock {\em CVPR}, 2016.

\bibitem{yang2016stacked}
Z.~Yang, X.~He, J.~Gao, L.~Deng, and A.~Smola.
\newblock Stacked attention networks for image question answering.
\newblock {\em CVPR}, 2016.

\bibitem{zhang2014review}
M.-L. Zhang and Z.-H. Zhou.
\newblock A review on multi-label learning algorithms.
\newblock {\em IEEE transactions on knowledge and data engineering},
  26(8):1819--1837, 2014.

\end{thebibliography}
}

\end{document}